\def\year{2020}\relax
\documentclass[letterpaper]{article} % DO NOT CHANGE THIS
\pdfoutput=1
\usepackage{aaai20}  % DO NOT CHANGE THIS
\usepackage{times}  % DO NOT CHANGE THIS
\usepackage{helvet} % DO NOT CHANGE THIS
\usepackage{courier}  % DO NOT CHANGE THIS
\usepackage[hyphens]{url}  % DO NOT CHANGE THIS
\usepackage{graphicx} % DO NOT CHANGE THIS
\usepackage{graphicx}  % DO NOT CHANGE THIS
\usepackage[sort&compress,square,numbers]{natbib}
\usepackage{url}
\usepackage{algorithm}
\usepackage{algorithmicx}
\usepackage{algpseudocode}
\usepackage{amsmath}
\usepackage{amsthm}

\newtheorem{definition}{Definition}
\newcommand{\rmem}[1]{\textrm{\emph{#1}}}
\algdef{SE}[DOWHILE]{Do}{doWhile}{\algorithmicdo}[1]{\algorithmicwhile\ #1}%

 \usepackage{color}

\title{
Faster and Safer Training by Embedding High-Level Knowledge into Deep Reinforcement Learning
}

\author{
Haodi Zhang\textsuperscript{\rm 1}，Zihang Gao\textsuperscript{\rm 1},Yi Zhou\textsuperscript{\rm 2}, Hao Zhang\textsuperscript{\rm 3}, Kaishun Wu\textsuperscript{\rm 1}, Fangzhen Lin\textsuperscript{\rm 4}\thanks{Corresponding Author}\\
\textsuperscript{\rm 1}College of Computer Science and Software Engineering, Shenzhen University\\
\textsuperscript{\rm 2}Shanghai Research Center for Brain Science and Brain-Inspired Intelligence/Zhangjiang Laboratory\\
\textsuperscript{\rm 3}Dorabot Inc.\\
\textsuperscript{\rm 4}Department of Comupter Science and Engineering, Hong Kong University of Science and Technology\\
}
 
\begin{document}

\maketitle

\begin{abstract}
%Deep reinforcement learning has been successfully admitted in many scenarios. However, the lack of interpretability brings obvious limit on intuitive understanding and explainable human interaction. Indeed end-to-end program save researchers from tedious implicit details in tasks, but the black-box-style approach also makes it difficult to embed highly abstracted human knowledge. In this paper, we introduce an framework that combines deep neural network with rules, which dynamically affect the training process. Our experiment on several games shows that, with the conditional supervision of these rules, deep Q-learning shows higher training efficiency. We also explore different supervising mechanism. The framework in this paper provides an intuitive way to embed human knowledge into the deep learning procedure. 

%Deep reinforcement learning has been successfully applied in many decision making scenarios. 
%However, the lack of interpretability prevents intuitive understanding and explainable intervention by human-level knowledge. 
%We propose a framework that embeds knowledge into deep reinforcement learning as knowledge. 
%The rules dynamically effect the training progress, and accelerate the learning. 
%The embedded knowledge not only improves learning efficiency, 
%but also prevents unnecessary or disastrous explorations at early stage of training. 
%Moreover, the modularity of the framework makes it straightforward to transfer high-level knowledge among similar tasks.

Deep reinforcement learning has been successfully used in many dynamic decision making domains, especially
those with very large state spaces.
However, it is also well-known that deep reinforcement learning can be very slow
and resource intensive. The resulting system is often
brittle and difficult to explain. 
In this paper, we attempt to address some of these problems by 
proposing a framework of Rule-interposing Learning (RIL) that 
embeds high level rules into the deep reinforcement learning.
With some good rules, this framework not only can accelerate the learning process,
but also keep it away from catastrophic explorations, thus 
making the system relatively stable even during the very early stage of training.
Moreover, given the rules are high level and easy to interpret, they can be
easily maintained, updated and shared with other similar tasks.

%The rules dynamically effect the training progress, and accelerate the learning. 
%The embedded knowledge not only improves learning efficiency, 
%but also prevents unnecessary or disastrous explorations at early stage of training. 

\end{abstract}

\section{Introduction}
\label{sec:intro}

Deep reinforcement learning \cite{mnih2013playing,mnih2015playing}
has been successfully used in many dynamic decision making domains, especially
those with very large state spaces.
Its showcase success stories include AlphaGo Zero
and for playing Atari video games.
However, like deep learning, it suffers from problems like being brittle
and not easily explainable. The training time is also often very long and
suffers from ``cold start'' - performing very badly at the beginning. 
Furthermore, for applications in robotics and critical decision support systems,
the lack of a guarantee that the system won't do anything disastrous is also
of concern.

%shown to be capable of successfully applied in many dynamic systems that require 
%intelligent decision making with a large state space.
%This end-to-end approach shows satisfactory performance and impressive generalizability in many domains such as Atari games. 
%However, like many other sub-symbolic approaches,  
%the black-box-style framework lacks of interpretability for its special structure and temporal abstraction. 
%Typically, when a deep neural network is introduced for some task, 
%there is no solid and explicit explanation for either the hyperparameter setting, or training progress, or trained models. 
%The shortcoming leads to the absence of an explainable uniform way to improve the network structure.

These problems are well-known and there has been much work on
addressing them.
There has been work on visualizing the behaviors of neural networks to help
human users understand them (e.g. \cite{Zahavy:2016:GBB:3045390.3045591,maaten2008visualizing}).
%These models and approaches provide interesting insight about behavior of deep neural networks. 
%But how the explanation can benefit the learning algorithm is a more important issue. 
There has also been work on using 
symbolic high-level planners
to guide neural networks learning process (e.g. \cite{DBLP:conf/icml/0001JADYD18,DBLP:conf/aaai/LyuYLG19}).
For instance, in \cite{DBLP:conf/icml/0001JADYD18} the symbolic module is responsible for high-level planning, 
while deep Q-Learning is deployed to accomplish each sub-task given by the high-level plan. 
There is also an interesting work \cite{DBLP:journals/ai/LeonettiIS16} that
first uses symbolic planning to come up with possible candidate solutions and then feeds these
candidate solutions to
a neural network for it to select the final solution.
Other proposals including Imitation Learning %and policy shaping approaches
that tries to learn directly from human
(e.g.
\cite{DBLP:conf/atal/SaundersSSE18,DBLP:conf/nips/GriffithSSIT13,DBLP:conf/ijcai/CederborgGIT15}).
%However, these approaches either require manual intervention such as sub-tasking or on-line oversight, 
%or presume computational power of symbolic planners.

In this paper, we propose a natural way to combine high-level symbolic rules with deep reinforcement learning.
These rules can be intuitive heuristics such as ``slow down when you approach the curve''. They can also be
safety rules like ``don't go too close to the cliff''. Our basic assumption is that these rules are
often easy to come by in many domains and are intuitive and easy to understand. They may not be complete and
detailed enough but should be useful to the agent during the learning process. To test our
hypothesis, we proposed a framework of rule-interposing learning (RIL) for combining rules and
deep Q-learning (DQN), one of the leading approaches to deep reinforcement learning. The idea is very
simple. During the reinforcement learning process, in addition to the current Q-values of possible actions,
consider also whether any of the rules is applicable, and if so, do it with certain probability that
depends on the stage of the learning and the type of the rules.

We have implemented our framework and tried it on some well-known domains such as
the Flappy bird, the Aircraft Shooting, the Breakout game, and the Grid World game.
The results were as we expected:
\begin{enumerate}
\item Good heuristic rules work as accelerators that make DQN learn faster.
\item Safety rules work as guards that make DQN learn more safely.
\end{enumerate}

It is worth emphasizing that
under the oversight of the safety rules, the network prevents ``disastrous'' explorations.
Therefore with appropriate safety rules, our framework can avoid ``cold start'' of DQN.
This is an important feature 
in those domains that are difficult to simulate and require on-site training, like what the ``I don't
want my robot do reinforcement learning in my kitchen'' slogan implies.

We also observed that in the end, some rules became ``obsolete'' as they became fully implemented by
the Q-networks. This again is not surprising but has good ramifications. It certainly increases the confidence
that one has on the learned network.
It should also help one to adapt
the learned network to similar domains as the rules are easily understandable and modified for the new domain.
This is somewhat related to 
transfer learning (e.g. \cite{DBLP:conf/icml/DaiYXY07,DBLP:conf/aaai/PanSYK08,DBLP:conf/pakdd/ZhangZY19,DBLP:conf/aaai/MoZLLY18,DBLP:journals/ai/ZhaoPY17}) but different.

The rest of the paper is organized as follows. In the next section, we describe our framework for integrating
rules into deep Q-learning. We then describe in details our experiments on
the following four games: Flappy bird, Aircraft Shooting, Breakout, and Grid World.
We next discuss some related work and then conclude the paper.

\section{Rule-interposing Learning}
\label{sec:framework}

Our rule-interposing learning (RIL) framework does not require sub-tasking, 
or any other manual intervention from human experts. 
Instead, we assume that each agent has some common-sense knowledge in the form of rules. 
Each rule consists of two parts: a precondition about the environment, 
and the recommended actions when the precondition is satisfied.
For instance, a rule in Flappy bird could be:
$$
\textrm{\emph{If the position of the bird is lower than a threshold, then flap.}}
$$
Unlike the human demonstration data, these rules are highly abstract and 
more easily described in natural language by human experts.
Moreover, it is also straightforward to represent in formal logic or action languages.
The rule above can be written as a first-order logic proposition:
$$
\rmem{lower(pos(bird), thresh) $\supset$ flap}.
$$
Alternatively it can also be written in action language $\cal BC$ \cite{DBLP:conf/ijcai/LeeLY13}:
$$
\rmem{lower(pos(bird), thresh) {\bf causes} flap}
$$
or in an Answer Set Programming language \cite{DBLP:journals/amai/Niemela99,DBLP:books/sp/99/MarekT99,DBLP:conf/aaai/Lifschitz08}:
$$
\rmem{flap $\;\leftarrow\;$ pos(bird)$<$thresh}.
$$
The rule above has only one deterministic action to suggest under the precondition.
To be more general, a given rule suggests conditionally a set of actions,
proclaiming that any action in the suggestion set is acceptable.
Simply by introducing a random function $rand$, the rule above can be represented as
$$
	\rmem{lower(pos(bird), thresh) $\supset$ rand(\{flap\})}
$$

Formally, for a given domain, the knowledge base $R$ consists of rules of form
$
	(\eta, \delta)
$
where $\eta$ is a first-order logic proposition indicating some environmental condition, 
and $\delta$ is a set of conditionally recommended actions, which is a subset of action space. 
For convenience, the two parts of a given rule $r\in R$ are written as functions in the rest of the paper,
denoted respectively by $\eta(r)$ and $\delta(r)$. 
Denote activation set of rule $r$ at timestamp $t$ as 
$$
	\alpha(r,t) =\left\{
		\begin{array}{cl} 
			\delta(r) & \textrm{if $\eta(r)$ is true at timestamp $t$}\\
			\emptyset & \textrm{otherwise.}
		\end{array}
	\right.
$$
The activation set $\alpha(r,t)$ contains all actions suggested by rule $r$ at time $t$,
and it is obviously also a subset of action space. 
The activation set of the entire knowledge base at time $t$ is defined as
the intersection of all non-empty activation sets of rules:
$$
	\alpha(R, t) = \bigcap\limits_{r\in R, \alpha(r,t)\not=\emptyset} \alpha(r,t).
$$
Especially, given a time stamp $t$, if $\alpha(r, t)=\emptyset$ for each rule $r\in R$,
it means that none of the rules applies in current situation. 
Therefore, DQNs should explore or select an action autonomously in this case.
At each timestamp $t$, there might be multiple non-empty activation sets.
\begin{definition}
	A knowledge base $R$ is \emph{consistent} at timestamp $t$ if 
	$|R|\leq 1$, or for any rules $r_i, r_j$ in $R$, 
		$$ \alpha(r_i,t)\not= \emptyset \land \alpha(r_j,t)\not= \emptyset \supset \alpha(r_i,t)\cap\alpha(r_j, t)\not= \emptyset.$$
\end{definition}
Ideally, the knowledge base should be always consistent at any time $t$.
If there are two different rules whose activation sets have no common suggested action, it is a conflict in the given knowledge.
In RIL, we simply ignore the rule set when there is conflict in the knowledge base.

\begin{figure}[t]
	\centering
	\includegraphics[width=0.9\columnwidth]{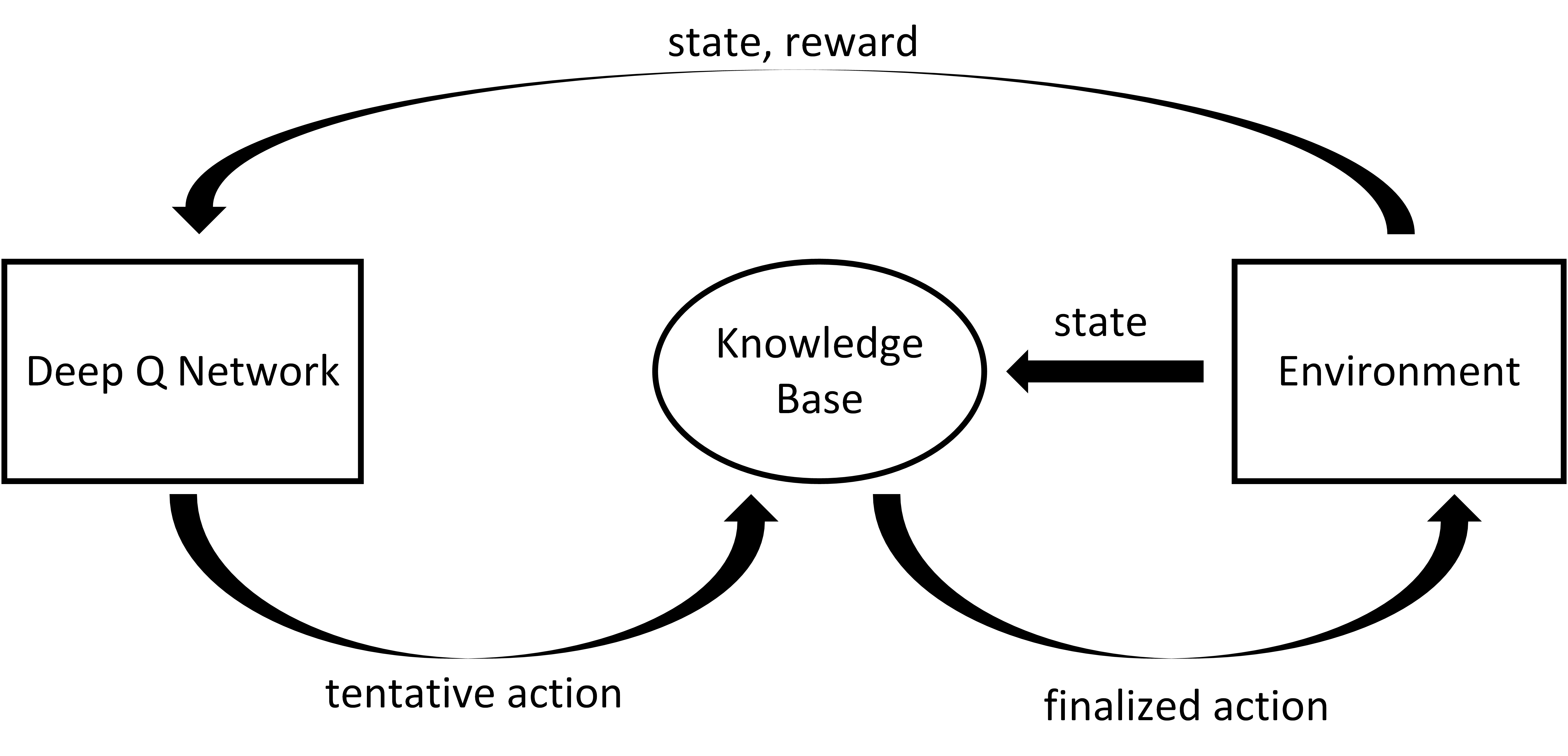}
	\caption{The architecture of RIL}
	\label{fig1}
\end{figure}

The knowledge base interacts with DQNs and prunes away unnecessary or unsafe explorations. 
Consequentially, the training of DQNs gains not only more efficiency, 
but also better performance even in very early stage of training. 
We here introduce two interposing mechanisms for the rules to illustrate the improvement.
The two mechanisms have a uniform representation. 
Before introducing the details of them, we give the architecture of RIL framework in Figure~\ref{fig1}.

\begin{algorithm*}
\caption{Rule-interposing Deep Q-Learning}
\label{alg:Framework} 
\begin{algorithmic}[1]
\Require Rule set $R$, initial interposing probability $p_0$, decay rate $\gamma$, 
	training round limit $(M, T)$, preprocession $\phi$, and exploration probability $\varepsilon$.
\State Initialize replay memory $D$ to capacity $N$
\State Initialize action-value function $Q$ to random weights $\theta$
\State Initialize target action-value function $Q^*$ to $\theta^*=\theta$
\While {not $episode=1$ to $M$}
 \State {Initialize sequence $s_1=\{x_1\}$ and preprocessed sequence $\phi_1=\phi(s_1)$}
 \For{$t=1$ to $T$}
  \State {With probability $\varepsilon$ set $a_{t}$ to a random action}
  \State {otherwise set $a_{t}=\max _{a} Q^*\left(\phi\left(s_{t}\right), a ; \theta\right)$}
  \State {Set $\alpha(R,t)=\bigcap\limits_{r\in R, \alpha(r,t)\not=\emptyset} \alpha(r,t)$}
  \If{$\alpha(R,t)$ is consistent and nonempty and $a_{t}\not\in \alpha(R,t)$}
   \State {With probability $P_t=p_0\cdot\gamma^{t}$, set $a_t$ to a random action in $\alpha(R,t)$}
  \EndIf
  \State {Execute action $a_{t}$ in emulator and observe reward $r_{t}$ and image $x_{t+1}$}
  \State {Set $s_{t+1}=s_{t}, a_{t}, x_{t+1}$ and preprocess $\phi_{t+1}=\phi\left(s_{t+1}\right)$}
  \State {Store transition $\left(\phi_{t}, a_{t}, r_{t}, \phi_{t+1}\right)$ in $D$}
  \State {Sample random minibatch of transitions $\left(\phi_{j}, a_{j}, r_{j}, \phi_{j+1}\right)$ from $D$}
%  \State {Set $y_{j}=\left\{\begin{array}{r}{r_{j}} {\;\,\,\;\qquad\qquad\qquad\textrm { if terminates at $j+1$}}  \\ 
%  			{r_{j}+\gamma \max _{a^{\prime}} Q\left(\phi_{j+1}, a^{\prime} ; \theta^*\right)} {\quad\textrm{otherwise}}\end{array}\right.$}
  \State {Set $y_{j}=\left\{\begin{array}{cc}{r_{j}} & {\textrm { if current episode terminates at $j+1$}}  \\ 
  			{r_{j}+\gamma \max _{a^{\prime}} Q^*\left(\phi_{j+1}, a^{\prime} ; \theta^*\right)} & {\textrm{otherwise}}\end{array}\right.$}
  \State Perform a gradient descent step on $(y_j-Q(\phi_j,a_j;\theta))^2$ with respect to the network parameters $\theta$
  \State Every $C$ steps reset $Q^*=Q$
 \EndFor
\EndWhile
\end{algorithmic}
\end{algorithm*}

The deep neural network gets screenshots, 
reward and termination signal form environment and outputs action for the agent to execute. 
The knowledge base checks the tentatively selected action and makes the decision.
how much power is authorized to the knowledge rules is decided by interposing mechanism, 
in term of setting effective probability dynamically.

To be specific, similar with original deep Q-learning, RIL first gets sequential screenshots, rewards and termination signals from the game environment. 
Then several continuous original game screenshots are processed to gray-scale image, and then capsuled with the rewards and actions taken as a sample. 
This sample is deemed as experience and restored in the replay memory with certain capacity $N$. 
When the replay memory is full, the out-of-time experience is popped up. 
In every training step, the model randomly gets a sample from the replay memory for training,
and calculates the predicted Q-value for every valid action:
\begin{eqnarray}
	Q^{*}(s, a)=E_{s^{\prime} \sim \mathcal{E}}\left[r+\gamma \max _{a^{\prime}} Q^{*}\left(s^{\prime}, a^{\prime}\right) | s, a\right]
\end{eqnarray}
The agent selects a random action with probability $\varepsilon$, otherwise select the action with maximal Q-value.
But unlike original DQN, before the execution of selected action, RIL passes the action into rule set.
The rule set maintain a pool of legal actions by updating $\alpha(r, t)$ for each rule in knowledge base $R$.
If the selected action violates the knowledge, RIL rejects the action under following probability
$$
	P_{t}=p_{0}\cdot\gamma^{t},
$$
where $p_0$ is a given initial probability, $\gamma$ is the decay rate, and $t$ is the timestamp.
After the rejection, a random legal action is selected to be executed.
Therefore, there are totally three sources of action to take: random exploration, the action with maximal Q-value, and the action derived by rules.
The framework decides the final action according to interposing mechanism.
Algorithm~\ref{alg:Framework} shows the exact learning process.
The algorithm employs original DQN to illustrate the effectiveness and generality of our approach.
Even with such a simple implementation, the rules improves the efficiency of deep learning dramatically.

In the following, we demonstrate RIL's performance under two rule-interposing schemes:

	\noindent\textbf{Acceleration rules:} the rules with probability $P_t=p_0\cdot\gamma^{t}$ where $0<\gamma <1$.
	Given existing knowledge about the task, some explorations are unnecessary and can be pruned.
	As a consequence, under the instruction of these rules as a priori, a DQN learns faster. 
	In early stage of the learning, the network is not yet well trained, so the rules might make much better decisions.
	With more rounds of training, the estimated Q-value of the action selected by DQN becomes more approximate.
	The rules are supposed to give more chance to DQN to decide. 
	Under the supervision of the rules, DQN gains information about the domain more efficiently.
	The improvement brought by introducing acceleration rules is well supported by our experiment.
	
	\noindent\textbf{Safety rules:} the rules with probability $P_t=p_0\cdot\gamma^{t}$ 
	where $p_0=1$ and $\gamma=1$.
	Obviously, in this case the rule will be always on, overseeing the training process. 
	Once the decision made by DQN is considered dangerous by the safety rules, 
	it'll be rejected and replace to a safe one given by knowledge base.
	In this way, the learning process is protected in a safe range in the environment.
	Our experiment shows that the performance of protected DQN is much better even in the early stage of learning.
	It is extremely useful for some learning tasks difficult to simulate, 
	enabling a cold boot and ensuring that the agent never touches those catastrophic explorations.

\section{Experiments}
\label{sec:exp}

\begin{figure}[t]
	\centering
	\includegraphics[width=\columnwidth]{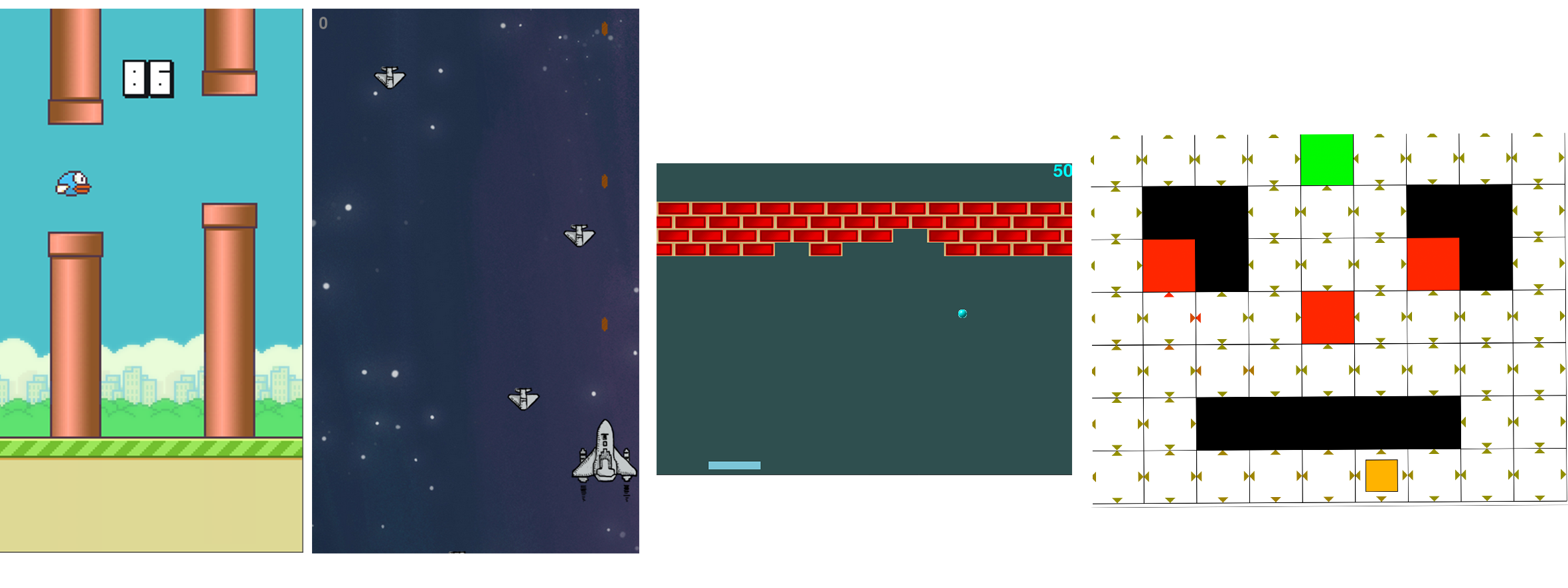}
	\caption{Screenshots from four games: (left-to-right) Flappy bird, Space war, Breakout and Grid world.}
	\label{fig2}
\end{figure}

In this section, RIL's performance is demonstrated by training DQNs to play several games, 
namely Flappy bird, Aircraft Shooting, Breakout and Grid World,
as shown in Figure~\ref{fig2}. The games are divided into two groups to demonstrate the two interposing schemes respectively.
To clarify, it is quite straightforward to integrate the two schemes into one algorithm. 
The knowledge base is divided into a safety set and a acceleration set, each with its own interposing probability,
as in Algorithm~\ref{alg:extend}. 
\begin{algorithm}
\caption{Extended RIL with Both Safety Rules and Acceleration Rules}
\label{alg:extend} 
\begin{algorithmic}[1]
\Require Safety rule set $R_s$, Acceleration rule set $R_a$, $p_0, \gamma, s, M, T, \phi$ and $\varepsilon$.
\State Initialize $D,Q$ and $Q^*$ as in Algorithm~\ref{alg:Framework}
\For{$episode=1$ to $M$}
 \State {Initialize sequence $s_1$ and $\phi_1$}
 \For{$t=1$ to $T$}
  \State {Set $\alpha(R_s,t)=\bigcap\limits_{r\in R_s, \alpha(r,t)\not=\emptyset} \alpha(r,t)$}
  \State {Set $\alpha(R_a,t)=\bigcap\limits_{r\in R_a, \alpha(r,t)\not=\emptyset} \alpha(r,t)$}
  \State {With probability $\varepsilon$ set $a_{tmp}$ to a random action}
  \State {otherwise set $a_{tmp}=\max _{a} Q^*\left(\phi\left(s_{t}\right), a ; \theta\right)$}
  \If {$\alpha(R_s,t)\not=\emptyset$ and $a_{tmp}\not\in \alpha(R,t)$}
   \If {$\alpha(R_a,t)\cap \alpha(R_s,t)\not= \emptyset$}
    \State {With probability $P_t=p_0\cdot\gamma^{t}$, set $a_t$ to a}
    \State { random action in $\alpha(R_a,t)\cap \alpha(R_s,t)$}
   \Else { Set $a_t$ to a random action in $\alpha(R_s,t)$}
   \EndIf
  \EndIf
  \State {Execute action $a_{t}$, observe $r_t$ and $x_{t+1}$}
  \State {Set the successor state, store the transition} 
  \State {Set $y_j$, perform gradient descent steps}
  \State {reset $Q^*=Q$}
 \EndFor
\EndFor
\end{algorithmic}
\end{algorithm}

In the experiment to introduce, the two interposing schemes 
work independently to demonstrate their respective roles.

All the four games share the same hyper-parameter setting,
including the network architecture (except for output layer, as the number of available actions varies in different games).
For each game, we compare our model RIL (DQN + knowledge) with the original DQN.
Notice that to have a fair comparison, the neural network implemented in RIL is also exactly the same with the baseline DQN. 
The network consists of three convolution layers, one hidden layer and the output layer. The first layer convolves the input image with an 8*8*4*32 kernel at a stride size of 4. The output is then put through a 2*2 max pooling layer. The second layer convolves with a 4*4*32*64 kernel at a stride of 2.  The third layer convolves with a 3*3*64*64 kernel at a stride of 1. The hidden layer consists of 256 fully connected ReLU nodes. For each game, we consider the agent's performance in two aspects. The main criterion is the average reward the agent gains in training episodes. The other one is the average Q-value which reflects the internal evaluation of current network.

\subsection{Acceleration rules}
The acceleration rules take effect with probability 
$$P_t=p_0\cdot\gamma^{t}$$ 
where $0<\gamma <1$.
In experiment we set $p_{0}$ to 1, and the decay rate $\gamma$ to 0.8.

\subsubsection{Flappy bird}

A bird manipulated by player attempts to fly across pairs of pipes, without hitting any. 
Two actions are available, namely $flap$ or doing nothing. 
By flapping the bird gets a temporary upwards acceleration, thus the bird can go up for a distance. 
If doing nothing, the bird will fall down due to the gravity. 
The bird gains reward by flying across pairs of pipes.
Once the bird hits a pipe or falls on the ground, the episode ends and loses some reward.

\begin{figure}[t]
	\centering
	\includegraphics[width=0.5\columnwidth]{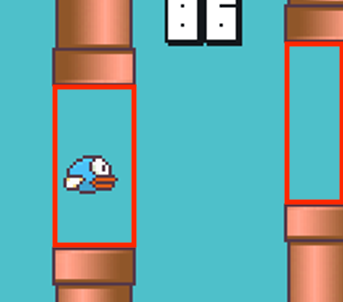}
	\caption{Effective regions of the rules in Flappy bird}
	\label{fig:precond}
\end{figure}

In this game, we use a rule set to tell the bird not to fly too high or too low, 
when it is flying across a pair of pipes. 
The rules only affect the training when the bird is flying in the red frames in Figure~\ref{fig:precond}.
Formally, knowledge base in Flappy bird $R_{fb}=\{r_1, r_2\}$, where $\eta(r_1)$ is 
$$
	\rmem{crossing}(p_u,p_l) \land \rmem{less}(\rmem{distance}(\rmem{bird}, p_u), \rmem{size}(\rmem{bird})),
$$
and $\delta(r_1)=\{flap\}$, and $\eta(r_2)$ is
$$
	\rmem{crossing}(p_u,p_l) \land \rmem{less}(\rmem{distance}(\rmem{bird}, p_l), \rmem{size}(\rmem{bird})),
$$
and $\delta(r_2)=\{null\}$,
where ($p_u$, $p_l$) is the pair of pipes that the bird is flying across.
Notice that $null$ represents doing nothing in the game, and $\delta(r_2)$ is different from empty set.

\begin{figure}[t]
	\centering
	\includegraphics[width=\columnwidth]{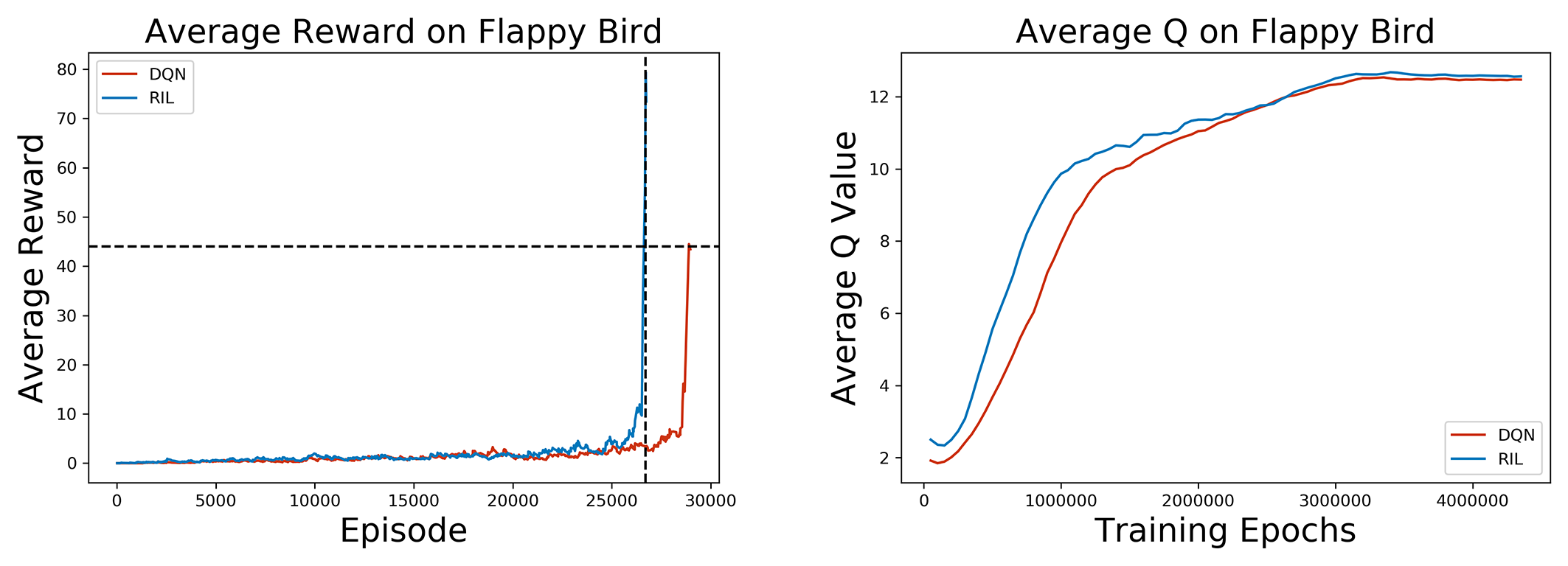}
	\caption{
	\textbf{Result of Acceleration Rules in Flappy bird.} 
	As the reward per episode keep increasing over time, we set a time limit of the training stage. 
	The reward per episode in left-hand side shows that, within the same training time,
	RIL gets better performance with fewer training episodes.
	The average Q-values in right-hand side also supports that the knowledge could accelerate the learning progress.
	}
	\label{fig3}
\end{figure}

The performance of the RIL framework in Flappy bird is shown in the Figure~\ref{fig3}. 
The plot of average reward of training episodes indicates obvious improvement on learning efficiency.
The conclusion is also supported by the average Q-value plot.

\subsubsection{Space war}

In this game, enemy planes appear randomly from the top of screen and dive vertically to the bottom. 
The agent controls a plane continuously shooting with a certain frequency. Each hit on the enemy plane gains some reward. 
If agent's plane collide with the enemy plane, the episode ends and loses the reward. 
The agent's airplane can only move horizontally, and the available actions are to move left and right. 

The rule set used in this game is a greedy strategy:
always move to the horizontally nearest enemy jet.
Formally, knowledge base in Space war is $R_{aw}=\{r_3, r_4\}$, where $\eta(r_3)$ is 
$$
	on\_left(nearest\_jet, agent)
$$
and 
$\delta(r_3)=\{move\_left\}$, and $\eta(r_4)$ is
$$
	on\_right(nearest\_jet, agent)
$$
and $\delta(r_4)=\{move\_left\}$.

As shown in Figure~\ref{fig5}, the learning process speeds up under the instruction of the knowledge base.

\begin{figure}[t]
	\centering
	\includegraphics[width=\columnwidth]{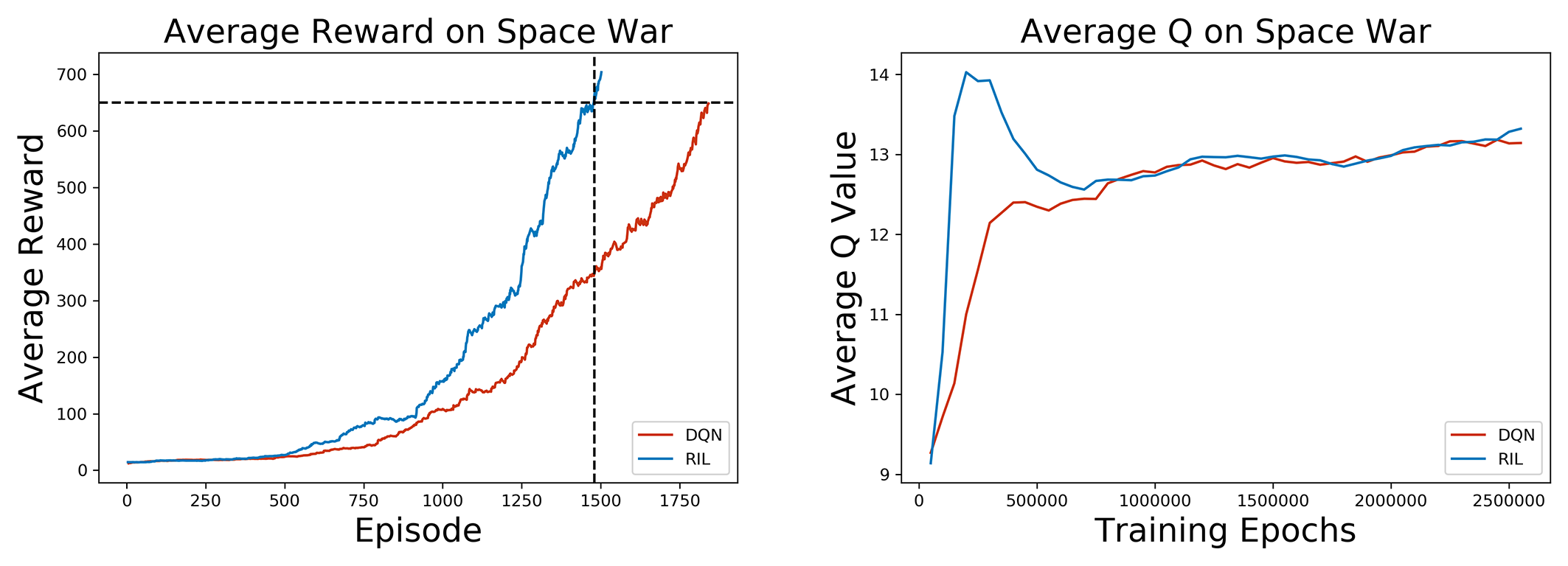}
	\caption{
	\textbf{Result of Acceleration Rules in Space war.} 
	A time limit is also set up.  The result shows that RIL drives DQN learn much faster than the original DQN.}
	\label{fig4}
\end{figure}

\noindent\textbf{Breakout} In this classic Atari game, we use following strategy:
if the ball is on the left-hand side of the paddle, then the paddle should move left,
the similar when it is on the right-hand side of the paddle.
Formally, the knowledge base for Breakout is $R_{bo}=\{r_5, r_6\}$,
where $\eta(r_5)$ is 
$on\_left(ball, paddle)$ and $\delta(r_5) = \{move\_left\}$,
and $\eta(r_6)=on\_right(ball, paddle)$ and $\delta(r_6) = \{move\_right\}$.
Figure~\ref{fig5} shows that with the strategy given by knowledge base $R_{bo}$, 
the DQN learns much faster.

\begin{figure}[t]
	\centering
	\includegraphics[width=\columnwidth]{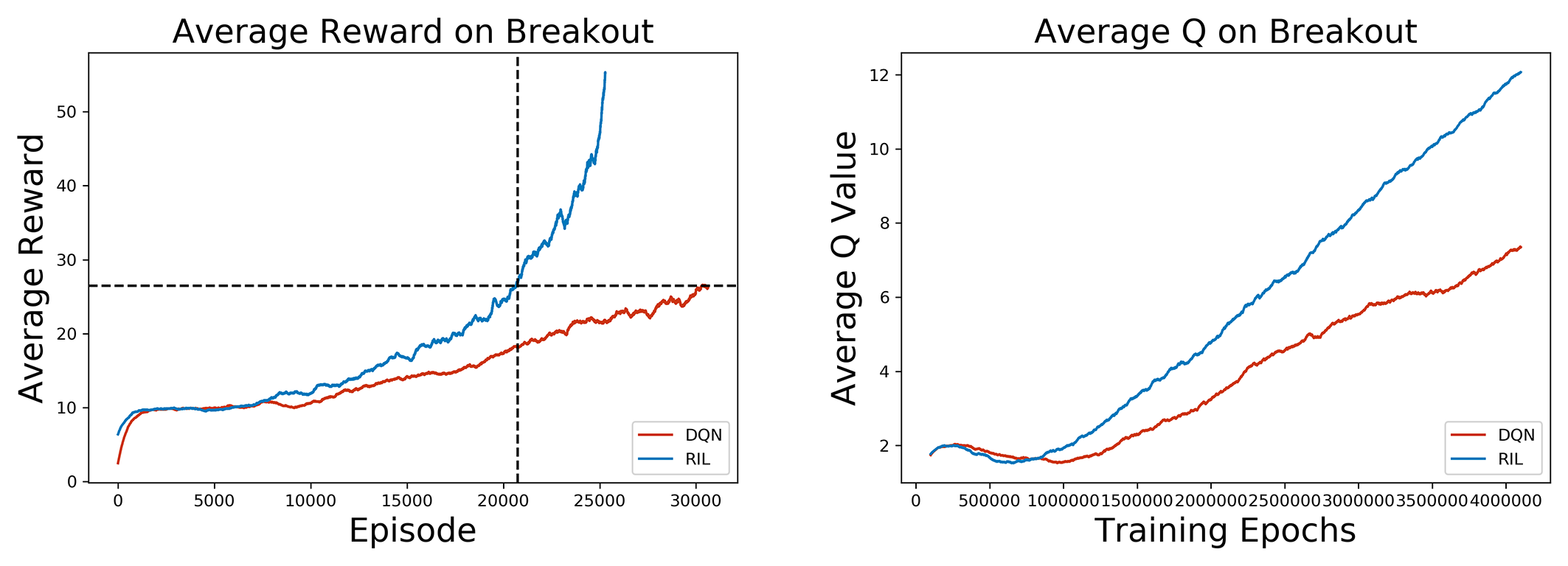}
	\caption{
	\textbf{Result of Acceleration Rules in Breakout.} Similar with result in other domains, DQN benefits from given knowledge and learns faster.
	}
	\label{fig5}
\end{figure}

Experiment result summarized in Table~\ref{tb:result} shows that acceleration rule set improves the learning efficiency. 

\begin{table}[]
\caption{Improvement by Accelerate Rule Set}
\label{tb:result}
%\begin{tabular}{ | p{1.7cm} | p{24mm} | p{28mm} |}
\begin{tabular}{ | c | c | c |}
\hline
\bf 			&	\bf Time saved 	&	\bf Reward improved  \\
\bf Games			&	\bf  with fixed reward	&	\bf with fixed time \\
\hline
\bf Flappy bird	&		20.00\%			&	7.67\%				\\
\hline
\bf Space war		&		7.93\%		&	81.82\%				\\
\hline
\bf Breakout		&		31.97\%		&	107.55\%				\\
\hline		
\end{tabular}
\end{table}

\subsection{safety rules}
In addition to acceleration rules, we can also have safety rules to ensure the safety of the agent during the training process. These rules are used to prevent the agent from doing actions that will cause some unrecoverable damage. Unlike the acceleration rules, these safety rules are enforced all the time during the
training process.
\subsubsection{Grid World}
\begin{figure}[t]
	\centering
	\includegraphics[width=0.9\columnwidth]{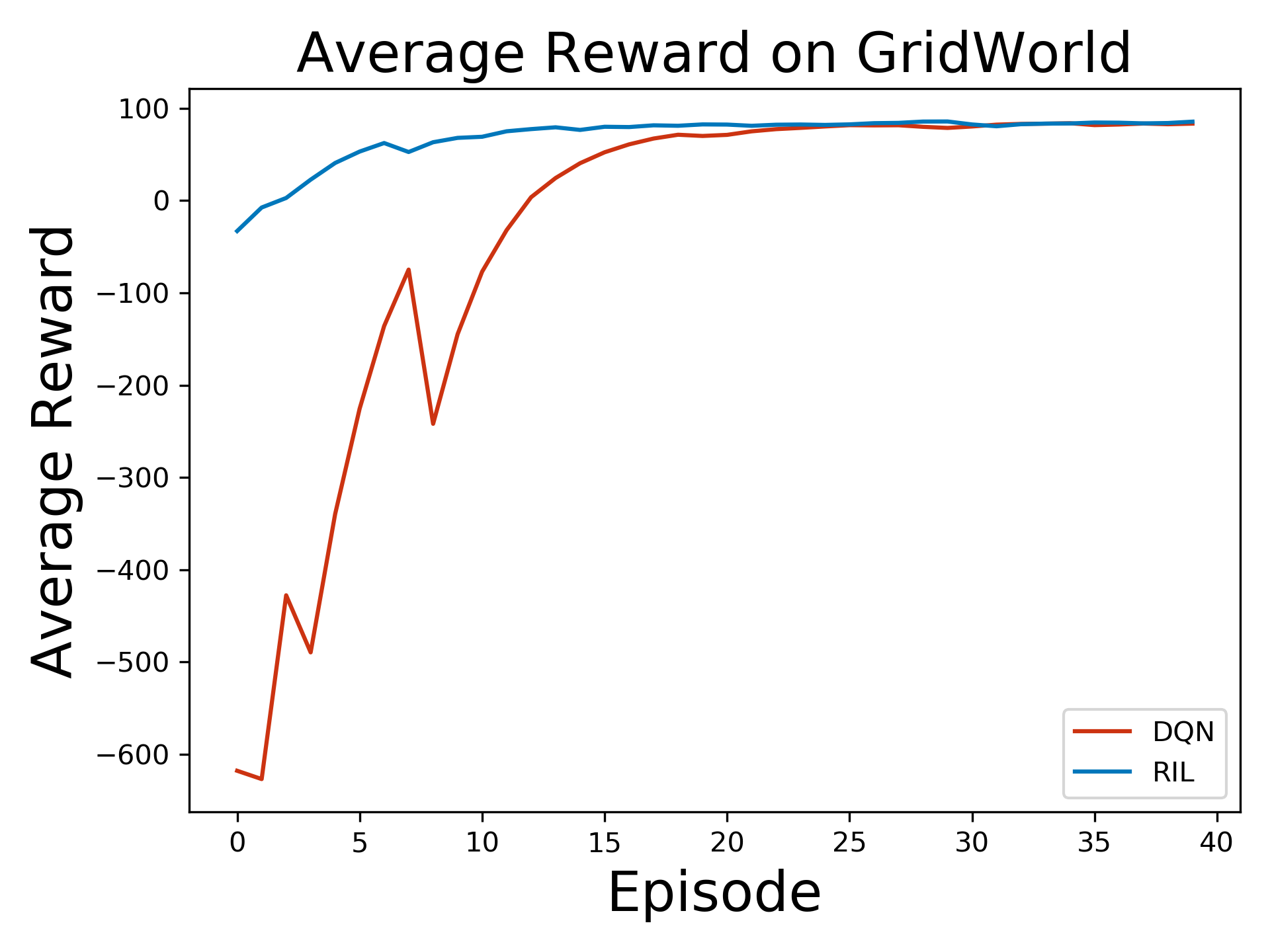}
	\caption{
	\textbf{Result of Safety Rules in Grid World.} 
	Safety rule successfully avoids catastrophic explorations, and the performance is much better in early stage of training.}
	\label{fig6}
\end{figure}

As in Figure~\ref{fig2}, the green grid is the destination, and black ones denote the walls that is unreachable, and the red one are traps. Once falling into the trap, the game ends and the agent gets a big penalty. The goal of agent is to find a shortest way to the destination without falling into a trap.
The agent get a negative reward of -1 for each move. If it falls into the traps, it'll get a penalty of 600. The reward of reaching the destination is 100.

In this game, we use the knowledge base $R_{gw}$ with a single safety rule $r_7$, which takes effect when the agent is in the neighbor of a trap,
where $\eta(r_7)$ is 
$$
near\_trap\land trap\_in(directions),
$$
and $\delta(r_7)$ is 
$${\cal A}-\{move(dir): dir\in directions\},$$ where $\cal A$ is the set of all actions.
The rule simply to forbid the agent to move into a trap.
This rule is a compulsory rule. 

The result in Figure~\ref{fig6} shows that ever in very early stage, the performance of NIL is much better than the Q learning algorithm. And during all the training process, the agent never gets into the traps, which ensures the safety of agent. Safety rules are especially useful when a cold boot is in need.

\subsection{knowledge sharing across domains}
\begin{figure}[t]
	\centering
	\includegraphics[width=0.9\columnwidth]{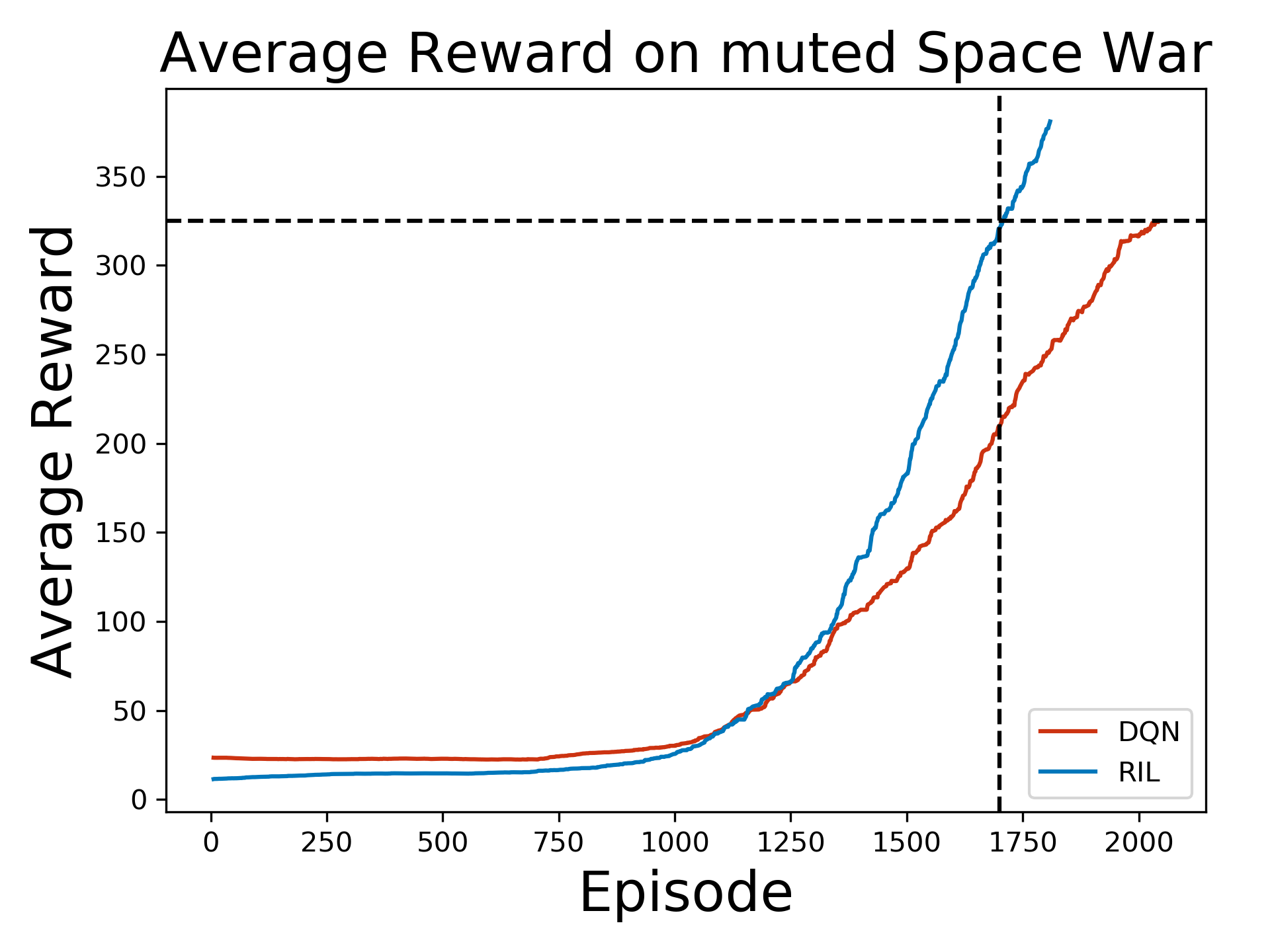}
	\caption{
	\textbf{Experiment result in muted Space war.} 
	With the same knowledge base and interposing scheme, the learning process improves in a similar way.}
	\label{fig7}
\end{figure}

Similar tasks can share the same rule set. We change the screen size of the Space war to define a new domain. 
For the DQN, it is completely a different game. But for human, the task is very similar to the original one. 
The same knowledge base and interposing scheme are used in the new learning task and the result is shown in Figure~\ref{fig7}.
With the same knowledge base deployed, the rules benefits DQN in a similar way as in Figure~\ref{fig4}.
A more interesting idea is to hire deep reinforcement learning to generate or update high-level rules, which forms a closed loop 
to share and transfer knowledge from one domain to another. We leave it in future work.

\section{Related Work}
\label{sec:related}

Since being proposed in 2013, deep Q-learning has gained much attention, and many variants have been proposed,
including Double DQN \cite{DBLP:conf/aaai/HasseltGS16}, Dueling DQN \cite{DBLP:conf/ICMLl/WangSHHLF16}, 
DRQN \cite{DBLP:conf/aaaifs/HausknechtS15}, prioritized DQN \cite{schaul2015prioritized}, 
bootstrapped DQN \cite{DBLP:conf/nips/OsbandBPR16}.  
These models differ in network structures, experience replay, $\varepsilon$-greedy or reward function
but have the same the core infrastructure.
They all have the same problems as the original deep Q-learning such as long training time and cold start.
None of them make use of high-level explainable domain knowledge.
For combining symbolic knowledge and deep reinforcement learning, we have mentioned
work that combines symbolic planning with deep Q-learning \cite{DBLP:conf/aaai/LyuYLG19}.
The basic idea is similar to hierarchical deep reinforcement learning 
\cite{DBLP:conf/nips/KulkarniNST16} that utilizes a meta-controller to learn to sequence subtasks defined on objects. 
Some applications can be found in \cite{DBLP:conf/icml/0001JADYD18,DBLP:conf/aaai/TesslerGZMM17,DBLP:conf/nips/KulkarniNST16,DBLP:conf/aaai/YinP17}.
Combination of symbolic planning and reinforcement learning is also implied in DARLING
\cite{DBLP:journals/ai/LeonettiIS16}, yet in very different way. Instead of giving the exact optimal solution, 
the planner provides several approximated solutions under some relaxation. 
These candidate solutions generated by the planner are then merged and passed to reinforcement learning module, 
to learn the finalized approximated policy. In DARLING, knowledge is represented as the candidate solutions, 
which commit to the relaxed constraints by symbolic rules.
It presumes the computational power of the symbolic planner.

Neural-symbolic systems \cite{DBLP:conf/acl/LiangBLFL17} construct a network from a given rule set to execute reasoning. \cite{DBLP:conf/acl/HuMLHX16} develops an iterative distillation method that transfers the structured information of logic rules into the weights of neural networks, which works good in NLP. Framework in \cite{xu2017semantic} encodes symbolic knowledge into the loss function of neural network. 
As for decision making filed, deep symbolic reinforcement learning (DSRL) \cite{garnelo2016towards} proposes an end-to-end reinforcement learning architecture comprising a neural back end and a symbolic front end, which takes the advantages of neural network and symbolic representation in some way. 
While in our work, the knowledge is represented by rules which can be shared, as an independent module from DQNs.

Another cluster of related research is Imitation Learning 
\cite{DBLP:conf/ijcai/ZhangTGBS19,DBLP:conf/ijcai/TorabiWS19a}, which enables agents to learn a policy through imitating
a human demonstrator's behaviors.
Standard imitation learning requires a large number of high-quality demonstration data,
which makes it not very practical. 
Imitation learning from observation requires only state demonstrations generated by the expert.
The data could be preferences or intervention from human experts.
However, human knowledge is usually not very clear for agent to directly learn, or furthermore share.
A typical work is HIRL \cite{DBLP:conf/atal/SaundersSSE18}, which is particularly related with our work. 
Like many other IL approaches, it requires a relatively heaven workload
of human intervention. The human expert needs to oversee the agent's decision at each time stamp
during training. Once some catastrophic action is generated by DQN, the human blocks it and 
manually take another safer action. The knowledge from human expert is not explicitly represented, before it is embedded implicitly into the black box.
Besides, the approach requires much more manual intervention. 
Some other policy shaping work 
\cite{DBLP:conf/nips/GriffithSSIT13,DBLP:conf/ijcai/CederborgGIT15,DBLP:conf/icml/MacGlashanHLPWR17} 
formulate human feedback as policy advice, and derive some algorithm
for converting that feedback into a policy. It is more reasonable and explainable, 
but still requires frequent human feedback during training.
Another main limitation of these models is that, knowledge update is quite expensive.
When some knowledge from human needs to be revised or corrected, 
the system has to apply a re-training with human intervention from the very beginning.
More work on this topic can be found in the survey \cite{DBLP:conf/ijcai/ZhangTGBS19}.
Works related to the safety rules can be found in \cite{mukadam2017tactical,alshiekh2018safe,fulton2018safe,garcia2015comprehensive,achiam2017constrained}

While the aforementioned work combines symbolic knowledge with deep reinforcement learning, our work
is unique in that it integrates rules directly into the learning process. One can even choose how
aggressively to apply these rules in the case of acceleration rules. Given they are high level, these rules
are easier to understand, maintain, and updated. The advantages of using these rules are that they
speed up the training process, can avoid cold start, and work as the starting point to explain the
resulting network.

\section{Conclusion}
\label{sec:con}
In this paper, we introduce a rule interposed learning framework for integrating
high-level rules and 
deep Q-learning.
As confirmed by our experiments, the interposed rules as domain knowledge benefit deep Q-learning in terms of data efficiency, 
exploration safety and high-level interpretability. 
We believe our approach is general enough to be used in other deep learning algorithms and we will
explore this in our future work.

\section*{Acknowledgment}

We would like to thank xxx, xxx, xxx and xxx (names omitted
here for blind review) for their good suggestions.
The work is supported by Fund xxx.

\bibliographystyle{aaai}
\bibliography{aaai20}

\end{document}